\documentclass[runningheads]{llncs}

\usepackage[mobile]{eccv}

\usepackage{eccvabbrv}

\usepackage{graphicx}
\usepackage{booktabs}

\usepackage[accsupp]{axessibility}  

\usepackage{hyperref}
\hypersetup{
  colorlinks=true,
  citecolor=eccvblue  
}

\usepackage{orcidlink}

\usepackage{fontawesome5}
\usepackage{graphicx}
\usepackage{svg}
\usepackage{algorithm}
\usepackage{algpseudocode}
\usepackage{amsmath}
\usepackage{setspace} 
\usepackage{wrapfig}
\usepackage{caption}
\usepackage{animate}
\usepackage{float}  

\newcommand{\method}{MeGAS\xspace}
\newcommand{\meltIcon}[1][1.0em]{%
  \raisebox{-0.3ex}{%
    \includegraphics[
      height=#1,
      keepaspectratio
    ]{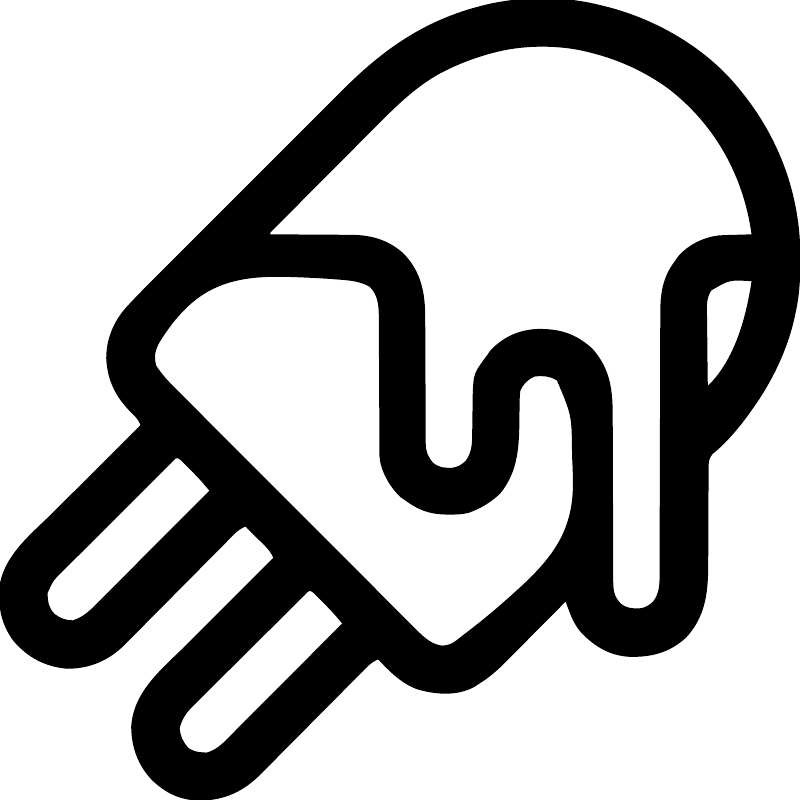}%
  }%
}

\definecolor{myPurple}{rgb}{0.4, .0, .8}
\definecolor{myGreen}{rgb}{0, 0.6, .3}
\definecolor{myRed}{rgb}{0.8, .2, .2}
\definecolor{myOrange}{rgb}{0.8, 0.45, 0.0}
\definecolor{myBlue}{rgb}{.0, .0, 1.0}
\definecolor{myBlue2}{rgb}{.0, 1.0, 1.0}
\definecolor{myBlack}{rgb}{.0, .0, 0.0}
\definecolor{darkmidnightblue}{rgb}{0.0, 0.2, 0.4}
\definecolor{MyGreen}{rgb}{0.02,0.5,0.02}

\newcommand{\zesong}[1]{\textcolor{myBlack}{#1}} 

\begin{document}

\title{%
\makebox[\textwidth][c]{%
\parbox{\dimexpr\textwidth+1.0em\relax}{\centering\sloppy
\method\hspace{-0.25em}\meltIcon[0.9em]:\ 
Thermomechanical Dynamic Gaussian Splatting for Thermophysical Scene Editing
}}
\vspace{-20pt}
}

\author{
Zesong Yang\inst{1*} \and
Yuanhang Lei\inst{1*} \and
Liyuan Cui\inst{1} \and
Yihang Chen\inst{1} \and
Jiaer Huang\inst{1} \and
Boming Zhao\inst{1} \and
Peter Yichen Chen\inst{2} \and
Hujun Bao\inst{1} \and
Zhaopeng Cui\inst{1}$^\dagger$
}

\authorrunning{F.~Author et al.}

\institute{
\vspace{-5pt}
$^1$State Key Laboratory of CAD\&CG, Zhejiang University \\
$^2$University of British Columbia \\
\vspace{1pt}
Project Page: \url{zju3dv.github.io/MeGAS}}

\titlerunning{MeGAS: Thermomechanical Dynamic Gaussian Splatting}
\authorrunning{Z. Yang et al.}

\maketitle

\setlength{\abovedisplayskip}{0.275 em} 
\setlength{\belowdisplayskip}{0.275 em} 

\begin{figure}[!htbp]
\centering
    \scriptsize
    \vspace{-2.5 em}
    \includegraphics[width=0.975\textwidth]{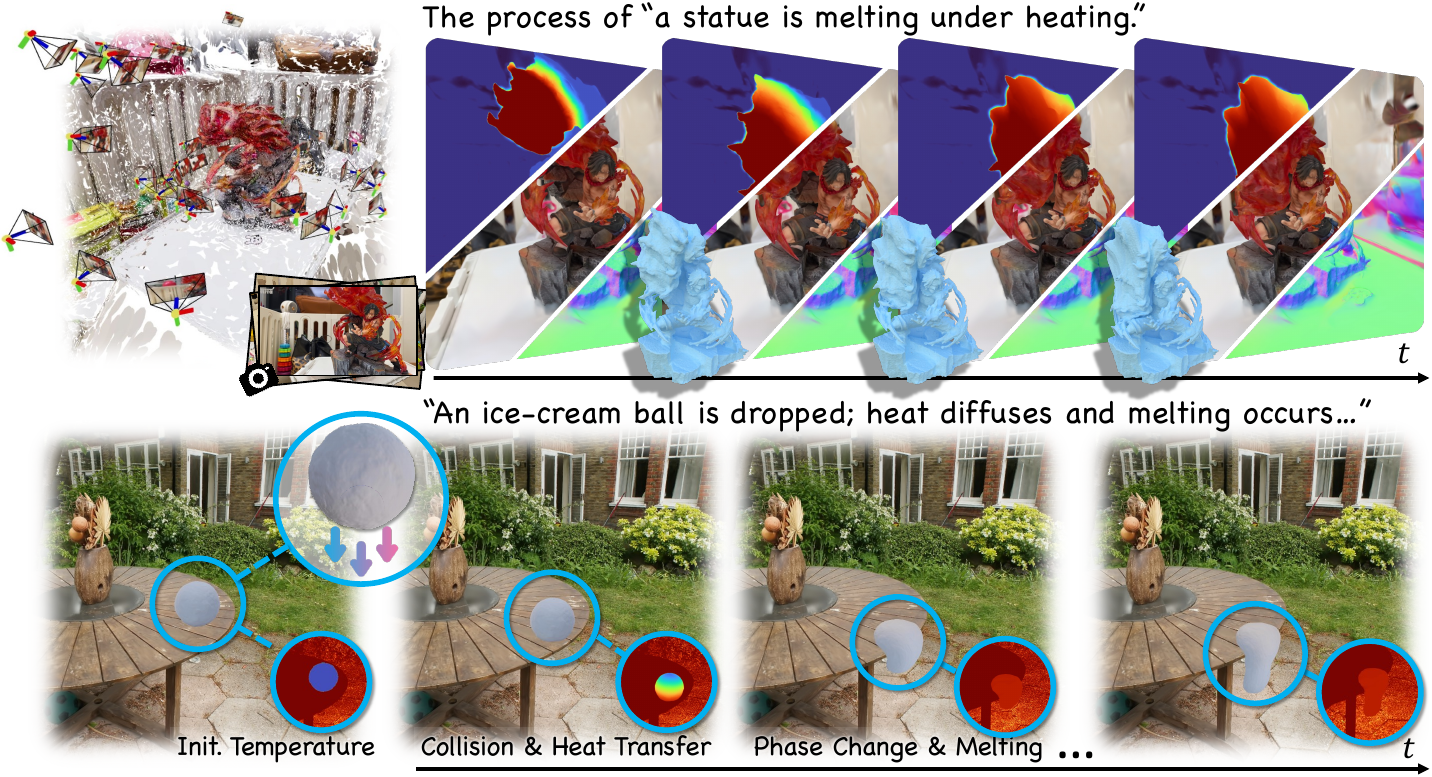}
    \caption{
    \textbf{Thermophysical editing and photorealistic rendering of real scenes. }
    \textbf{\method} integrates thermomechanical dynamics into 3D Gaussian Splatting, enabling physically grounded editing with temporally evolving temperature fields, while maintaining smooth and continuous geometry under large deformations.
    }
    \label{fig:teaser}
\end{figure}

\let\oldthefootnote\thefootnote
\renewcommand{\thefootnote}{}
\footnotetext{* Equal contribution. $\dagger$ Corresponding Author.}
\let\thefootnote\oldthefootnote

\vspace{-1.5 em}
\begin{abstract}
Recent advances integrate physically grounded Newtonian dynamics with neural rendering frameworks, narrowing the gap between photorealistic scene reconstruction and physics-based animation. However, existing approaches focus on mechanically driven dynamics while neglecting temperature, a fundamental yet invisible physical factor 
underlying phenomena such as melting, solidification, and other thermomechanical processes.
In this paper, we propose \textbf{\method}, a novel framework that incorporates thermomechanical phase-change dynamics into 3D Gaussian Splatting (3DGS). 
Specifically, we propose a new thermomechanical dynamic Gaussian Splatting representation that augments 3DGS with temperature attributes and employs a heat advection-diffusion solver with MPM dynamics incorporating phase transitions, enabling physically plausible and visually realistic synthesis of \zesong{thermophysical phenomena}. 
Furthermore, a new topology-adaptive Gaussian rendering strategy
is proposed to mitigate cracking and floaters under extreme deformation.
Extensive experiments demonstrate that \method produces physically consistent thermomechanical behavior while maintaining high-fidelity photorealistic rendering, advancing toward physics-integrated world models. 
\vspace{-1.0 em}
\vspace{-1.0 em}
\end{abstract}
\vspace{-1.0 em}
\section{Introduction}
\label{sec:intro}
\vspace{-0.5 em}

Recent advances in neural scene representations, such as Neural Radiance Fields (NeRF)~\cite{mildenhall2021nerf} and 3D Gaussian Splatting (3DGS)~\cite{kerbl3Dgaussians}, have 
significantly advanced
photorealistic scene reconstruction and rendering from real-world imagery, 
substantially promoting the development toward visually faithful digital twins and world models.
While these approaches achieve impressive visual fidelity, a practical open-world simulator requires not only visual realism but also the controllable synthesis of physically plausible phenomena that faithfully adhere to the physical laws.
To this end, recent research has integrated physical simulators into neural rendering, enabling the simulation of diverse phenomena such as elastic deformation~\cite{li2023pac,xie2024physgaussian,feng2024pie,zhong2024reconstruction,wang2025decoupledgaussian,shen2025enliveninggs}, fluid animation~\cite{feng2025gaussian}, and weather effects~\cite{li2023climatenerf,dai2025rainygs, shen2026fierygs}.

However, most existing approaches primarily focus on mechanically driven dynamics, while thermomechanical dynamics remain largely unexplored. In the physical world, temperature, an invisible yet fundamental quantity, governs a broad spectrum of visually perceivable phenomena, ranging from the melting of ice cream to the solidification of lava. 
Although several works have incorporated thermal information, they typically treat temperature as an auxiliary sensing modality for multispectral reconstruction~\cite{ye2024thermal,lin2024thermalnerf,ozer2024exploring,lu2024thermalgaussian} or estimate spatio-temporal temperature fields restricted in static scenes~\cite{chen2024thermal3d, yang2025ntr}.
Nevertheless, since thermomechanical dynamics are among some of the most intricate natural phenomena, integrating thermomechanical effects into neural rendering frameworks remains highly challenging.
\textbf{First}, thermomechanical dynamics necessitate not only the enforcement of momentum conservation but also the coupled modeling of heat transfer and phase transitions to produce physically plausible outcomes.
\textbf{Second}, \zesong{thermomechanical coupling (e.g., in melting)} introduces extreme deformations and topological changes, posing a significant challenge for neural rendering representations that typically assume continuous or smoothly deforming geometry.
Such extreme deformations cause sharp and spiky artifacts in splatting-based rendering, and the object surface tends to crack and expose interior regions, further breaking the geometric continuity for stable rendering. 
Consequently, a straightforward combination of existing Gaussian pipelines with thermomechanical dynamics fails to ensure visually stable results.



To address these challenges, we propose \textbf{\method}, a physics-integrated neural rendering framework that integrates thermomechanical dynamics with 3D Gaussian Splatting, 
\zesong{enabling controllable and thermophysically grounded editing of real scenes as shown in Fig.~\ref{fig:teaser}.} 
To this end, we first propose a new thermomechanical dynamic Gaussian Splatting representation that augments each Gaussian with a per-splat temperature attribute and supports direct rendering of a consistent thermal field within the standard 3DGS formulation. Temperatures can be flexibly initialized and manipulated through user-specified heat sources, providing intuitive and physically grounded editing control. 
To evolve this thermal field, we adopt a 
heat advection-diffusion solver
defined on a simulation grid. The solver propagates heat diffusion through the reconstructed scene, yielding a smooth, physical temperature distribution.
We further couple this thermal field to a phase-change–aware 
constitutive model switching,
enabling particle-level melting behavior: once a particle’s temperature exceeds the melting threshold, its constitutive model transitions from an elastic solid to a viscoplastic formulation, producing physically grounded and spatially localized melting dynamics.

Since melting induces large-scale deformation and surface topology changes, directly applying such extreme motion to vanilla or PhysGaussian-style~\cite{xie2024physgaussian} 3DGS leads to severe artifacts such as cracking, floaters, and interior exposure. 
We therefore design a novel Topology-Adaptive Gaussian Rendering strategy to ensure stable rendering throughout the melting process. Specifically, during reconstruction, we regularize Gaussian anisotropy to suppress excessively elongated kernels, enforce volumetric consistency, and prune internal Gaussians that would otherwise become visible under deformation. During animation, we perform an online surface-aware refinement that combines adaptive densification with local Gaussian kernel refitting guided by the evolving surface, which seals emerging cracks and maintains a coherent splat distribution as the object undergoes drastic thermomechanical change.

Our main contributions are summarized as follows:
\vspace{-0.5 em}
\begin{itemize}
\item We introduce \textbf{\method}, a novel framework that integrates thermomechanical phase-change dynamics  
into 3D Gaussian Splatting (3DGS), enabling physically plausible and photorealistic synthesis of \zesong{thermophysical} phenomena from real-world imagery.
\item We propose a new thermomechanical dynamic Gaussian Splatting representation that jointly models appearance and a controllable thermal field, while integrating heat advection-diffusion and temperature-controlled constitutive model switching, 
enabling melting dynamics controlled by phase change.
\item We 
develop a novel topology-adaptive Gaussian Rendering strategy with anisotropy regularization, internal-free filtering, and online surface-aware refinement to effectively mitigate cracking, floaters, and interior exposure under extreme deformation.
\item Extensive experiments on real and synthetic scenes demonstrate that our method achieves thermophysically consistent scene editing while preserving photorealistic rendering quality.

\end{itemize}

\begin{figure}[t] 
    \centering
    \scriptsize
    \includegraphics[width=\textwidth]{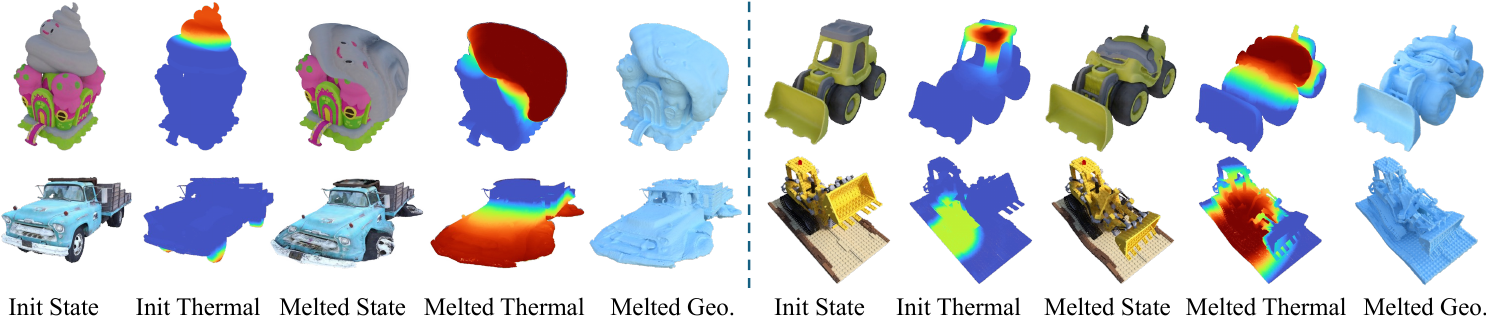}
    \caption{
    \textbf{Thermo-geometry evolution with \method.} 
    Under user-specified heating source, \method performs thermomechanically consistent phase-change simulation, steering the evolution of temperature fields and shape and delivering controllable, targeted melting without compromising geometric smoothness.
    }
    \vspace{-0.5 em}
    \label{fig:our_process}
\end{figure}

\section{Related Work}
\label{sec:relatedwork}

\noindent\textbf{Physics-Integrated Neural Rendering.}
Recent studies have integrated physical simulators into neural rendering frameworks, bridging photorealistic reconstruction with physically grounded realism. Early works~\cite{li2023pac,feng2024pie} embed continuum and elastodynamic models into NeRF, enabling geometry-consistent simulation of elastic and interactive deformations.
Subsequent researches~\cite{jiang2024vr, xie2024physgaussian,zhong2024reconstruction,wang2025decoupledgaussian,qiu2024feature,vasile2025asdiffmpm, leidiffwind, linomniphysgs} adapt 3D Gaussian Splatting (3DGS) to model mechanical dynamics while preserving rendering realism.
Beyond elasticity, \cite{feng2025gaussian} unifies position-based dynamics with 3DGS for fluid animation, and \cite{dai2025rainygs,li2023climatenerf, shen2026fierygs} synthesize environmental effects such as rainfall, snowfall and fire within neural rendering frameworks.
Despite these advances, existing approaches primarily focus on \emph{mechanically driven} or \emph{environmentally induced} phenomena.
In contrast, many natural phenomena are \emph{thermally driven}, where invisible temperature variations govern visually perceivable behaviors such as softening and melting.
Our work makes the first exploration of thermomechanical phase-change dynamics in neural rendering, addressing this underexplored yet fundamental dimension of physics integration.

\noindent\textbf{Thermal-Integrated 3DGS.}
Recent works extend neural rendering to thermal modalities by incorporating infrared imagery for multimodal scene reconstruction.
\cite{ye2024thermal,lin2024thermalnerf,ozer2024exploring,lu2024thermalgaussian} leverage paired RGB–thermal inputs to improve novel view synthesis via multispectral supervision, yet they rely purely on image-level supervision without explicit thermodynamic modeling.
\cite{chen2024thermal3d, yang2025ntr, wang2026etgs} couple 3DGS with simplified heat diffusion to estimate spatiotemporal temperature variations, achieving dynamic thermal field reconstruction but still limited to static scenes without considering more complex thermodynamic phenomena.
In contrast, we integrate a physically grounded thermomechanical model into 3DGS, 
capturing the coupled heat transfer and deformation dynamics for realistic melting synthesis.
\section{Preliminaries}
\label{sec:preliminary}
\noindent\textbf{3D Gaussian Splatting.}
Given posed RGB image sequences as input, we first reconstruct the scene with 3D Gaussian Splatting (3DGS)~\cite{kerbl3Dgaussians}.
3DGS represents the scene with a set of explicit 3D Gaussians, each parameterized by a center $\mu$ and a full covariance matrix $\Sigma$:
\begin{equation}
G(x) = e^{-\frac{1}{2}(x-\mu)^{\top}\Sigma^{-1}(x-\mu)}.
\end{equation}
For differentiable rendering optimization, $\Sigma$ is decomposed into a scaling matrix $S$ and rotation matrix $R$, i.e., $\Sigma = R S S^T R^T$,
where $S$ and $R$ are stored by a 3D scale vector $s$ and a quaternion $q$ respectively.
Given a viewing transformation $W$, the 3D Gaussians are projected to the image plane, and we obtain the 2D covariance matrix $\Sigma'$ and 2D center location $\mu'$ as: 
\begin{equation}
        \Sigma' = JW\Sigma W^{T}J^{T}, \quad
        \mu' = JW\mu,
\end{equation}
where $J$ is the Jacobian of the affine approximation of the projective
transformation. 
Then we can use the point-based volume rendering to render the color $C$ of each pixel with $N$ ordered 3D Gaussians:
\begin{equation}
    C = \sum_{i \in N}T_ic_i\alpha_i,
    \quad
    T_i = {\textstyle \prod_{j=1}^{i-1} (1 - \alpha_j)},
\label{eq:volume_render}
\end{equation}
with alpha $\alpha_i$ defined as: 
\vspace{-0.25 em}
\begin{equation}
\alpha_i = o_i e^{-\frac{1}{2}(x - \mu'_i)^{\top} \Sigma'^{-1}_i (x - \mu'_i)}.
\end{equation}
Here, $o_i$ and $c_i$ denote the learned opacity and color parameterized by spherical 
\noindent harmonics for each Gaussian.
Besides, we define the normal $\text{n}_i$ as the shortest axis direction.
\\
\noindent\textbf{Constitutive Governing Equations.}
We use the hybrid Eulerian-Lagrangian framework, Material Point Method (MPM)~\cite{jiang2015affine,hu2018moving}, to simulate continuum dynamics. The governing equations follow the conservation of mass and momentum:
\vspace{-0.5 em}
\begin{equation}
\frac{D\rho}{Dt} + \rho \nabla \cdot \mathbf{v} = 0, \quad
\rho \frac{D\mathbf{v}}{Dt} = \nabla \cdot \boldsymbol{\sigma} + \mathbf{f}^{ext},
\end{equation}
where $\rho$, $\mathbf{v}$, and $\boldsymbol{\sigma}$ denote the density, velocity, and Cauchy stress tensor respectively, and $\mathbf{f}^{ext}$ is the external force.
The stress tensor $\boldsymbol{\sigma}(\mathbf{x}, t)$ is derived from the hyperelastic constitutive model as:
\begin{equation}
\boldsymbol{\sigma}(\mathbf{x}, t) =
\frac{1}{\det(\mathbf{F})}
\frac{\partial \Psi}{\partial \mathbf{F}^E}
(\mathbf{F}^E)^{T},
\end{equation}
where $\Psi: \mathbb{R}^{3\times3} \rightarrow \mathbb{R}$ is the hyperelastic energy density function, and $\mathbf{F}^E \in \mathbb{R}^{3\times3}$ represents the elastic part of the deformation gradient $\mathbf{F}$. By specifying different formulations of $\Psi$, various constitutive models can be incorporated to characterize distinct material behaviors.

\begin{figure*}[t] 
    \centering
    \scriptsize
    \includegraphics[width=\textwidth]{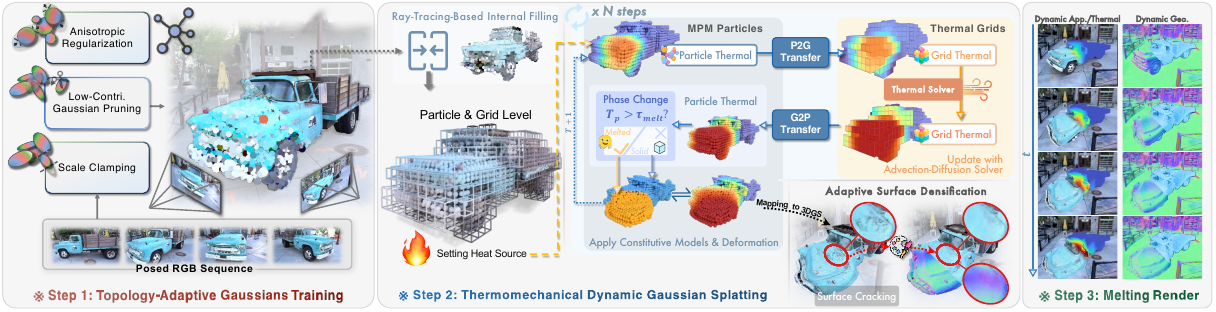}
    \caption{
    \textbf{System overview.}
    Starting from posed RGB sequences, we reconstruct the scene with 3DGS. To mitigate floaters and interior artifacts under extreme deformation, we introduce an interior-free uniform-Gaussian regularization that suppresses anisotropy and prunes unsupported interior splats. We then provide volumetric support via ray-tracing–based internal filling. Next, we augment 3DGS with per-Gaussian temperature and, driven by user-specified heat sources, \method updates temperatures using a grid-based heat advection-diffusion solver within an MPM simulator, and performs temperature-dependent phase-aware constitutive switching. Deformations are mapped back to 3DGS, and we apply adaptive surface densification to handle topology change and seal cracks.
    Finally, we synthesize physically plausible thermomechanical dynamics with photorealistic appearance and temporally smooth geometry.
    }
    \vspace{1.5 em}
    \label{fig:pipeline}
\end{figure*}

\section{Methods}
\label{sec:methods}

As shown in Fig.~\ref{fig:pipeline}, given posed RGB image sequences as input, we first reconstruct the scene with 3D Gaussian Splatting (3DGS)~\cite{kerbl3Dgaussians}.
Since melting serves as a representative and visually salient stress test of thermomechanical coupling, we adopt it as the primary instantiation to present our framework.
To simulate temperature-driven editing, we propose a thermomechanical dynamic Gaussian Splatting representation. Each Gaussian is augmented with a temperature attribute and we couple 3DGS to an MPM simulator with a heat advection–diffusion solver and a phase-change–aware constitutive model switching, enabling melting dynamics at the particle level. 
To maintain stable rendering under extreme deformation, 
we introduce a topology-adaptive Gaussian rendering strategy to obtain a uniform-distributed, volume-consistent, floater-free 3DGS, and incorporate a Ray-Tracing–based internal filling strategy to provide volumetric support for MPM. Finally, to prevent surface cracking during large deformations, we employ implicit-surface–guided densification that adaptively fills hollow regions with smoothly aligned Gaussian splats.

{
\setlength{\textfloatsep}{4pt plus 1pt minus 1pt} 
\setlength{\floatsep}{4pt plus 1pt minus 1pt}
\setlength{\intextsep}{4pt plus 1pt minus 1pt}

\makeatletter
\setlength{\@fptop}{0pt}  
\makeatother

\vspace{-1.0 em}
\subsection{Thermomechanical Dynamic Gaussian Splats}
\label{sec:method:simulation}

We augment each Gaussian with a temperature attribute $T_i \in \mathbb{R}$, and the corresponding temperature rendering of each pixel is obtained following Eq.~\ref{eq:volume_render}:
\begin{equation}
    \hat{T} = \sum_{i \in N} w_i\, T_i,
\label{eq:thermal_render}
\end{equation}
where \( w_i \) are the Gaussian weights along the viewing ray.

\noindent\textbf{Thermal Advection-Diffusion Solver.} 
To drive melting, we require a physically grounded evolution of the temperature field. Instead of heuristically editing, we solve heat diffusion on a simulation grid.
Specifically, we model heat transport with the advection-diffusion equation (ADE)~\cite{ashgriz2002introduction}:
\begin{equation}
    \frac{\partial T}{\partial t} + \mathbf{u} \cdot \nabla T = \alpha \nabla^2 T + S,
\label{eq:thermal_ADE}
\end{equation}
where \(T\) is the temperature field, \(\mathbf{u}\) is the thermal velocity field, \(\alpha\) is the thermal diffusivity, and \(S\) represents external heat sources.
The velocity \(\mathbf{u}\) is modeled by the Navier-Stokes equations (NSE)~\cite{batchelor2000introduction}:
\begin{equation}
    \rho \left( \frac{\partial \mathbf{u}}{\partial t} + \mathbf{u} \cdot \nabla \mathbf{u} \right) = -\nabla p + \mu \nabla^2 \mathbf{u} + \mathbf{f},
\label{eq:nse}
\end{equation}
with density \(\rho\), pressure \(p\), dynamic viscosity \(\mu\), and external forces \(\mathbf{f}\).

To advance the coupled system \eqref{eq:thermal_ADE}--\eqref{eq:nse}, we adopt the Lattice Boltzmann Method (LBM) that solves the flow and temperature fields and naturally captures both diffusion and advection~\cite{succi2001lattice}.
Additionally, thermal buoyancy is incorporated via the Boussinesq approximation as a body force~\cite{guo2002discrete}:
\begin{equation}
    \mathbf{F}_{b} \;=\; \beta\,\rho_{0}\,\bigl(T - T_{0}\bigr)\,\mathbf{g},
\label{eq:buoyancy}
\end{equation}
where \(\beta\) is the thermal expansion coefficient, \(\rho_{0}\) the reference density at temperature \(T_{0}\), and \(\mathbf{g}\) gravity. We add \(\mathbf{F}_{b}\) to \(\mathbf{f}\) in \eqref{eq:nse} for correction, avoiding explicit density updates while enabling accurate temperature-driven convection. 
See \textbf{supplementary material} for more details.

{
\makeatletter
\setlength{\@fptop}{0pt}      
\makeatother

\setlength{\textfloatsep}{0pt plus 1pt minus 1pt}
\setlength{\intextsep}{0pt plus 1pt minus 1pt}
\setlength{\floatsep}{6pt plus 1pt minus 1pt}

\begin{algorithm}[!t]
\caption{Thermomechanical MPM for Thermophysical Dynamics}
\label{alg:mpm}
\begin{algorithmic}[1]
\setstretch{1.05}
\renewcommand{\algorithmiccomment}[1]{\hfill$\triangleright$ #1}

\State \textbf{Notation}: 
$p, i$: particle \& grid indices; 
$m$: mass; 
$x$: position;
$w_{ip}$: P2G2P weight function;
$T$: temperatures; 
$\mathbf{F}^E$: elastic deformation gradient;
$\tau$: Kirchhoff stress;

\While{time\_step $<$ total\_steps}

    \ForAll{grid $i$} 
    \Comment{\textbf{Particle-to-Grid Transfer}}
        \State Transfer Mass \& Momentum to Grid via APIC
        \State {$m_i T_i \gets \sum_p w_{ip} m_p
        \big( T_p + (\mathbf{x}_i - \mathbf{x}_p) \cdot \nabla T_p \big)$} 
        \Comment{Transfer Thermal Field}
    \EndFor

    \ForAll{grid $i$}
    \Comment{\textbf{Grid Update}}
        \State Update Grid Velocities
        \State \textbf{{Update Thermal Boundary}}
        \State  {$T_i = \mathrm{LBM\_Update}(T_i)$ \Comment{\textbf{Thermal Evolution}}}
    \EndFor

    \ForAll{particle $p$}
    \Comment{\textbf{Grid-to-Particle Transfer}}
        \State Interpolate Velocity from Grid
        \State Update Particle Mechanical Properties

        \State  {$T_p \gets \sum_i w_{ip}T_i $} 
        \State  {$\nabla T_p \gets \sum_i T_i (\nabla w_{ip})^{T}$}
        \If{ {$T_p > T_{\mathrm{melt}}$}} 
            \Comment{ {\textbf{Phase Transition Update}}}
            \State  {\{$\tau_p, \mathbf{F}_p^{E}\} \gets \tau_{\text{liquid}}(\mathbf{F}_p^{E})$}
            \Comment{ {\textbf{Switch Material}}}
        \Else
            \State  {\{$\tau_p, \mathbf{F}_p^{E}\} \gets \tau_{\text{solid}}(\mathbf{F}_p^{E})$}
        \EndIf
    \EndFor
\EndWhile

\end{algorithmic}

\end{algorithm}

}

\noindent\textbf{Thermal-Driven Material Point Method.}
While we have defined how temperature is represented and evolved, building upon the standard MPM framework~\cite{hu2018moving,jiang2015affine,zong2023neural}, 
we extend the particle and grid states with thermal attributes, enabling temperature-dependent material behavior. 
During each time step, in addition to the standard MPM updates of mechanical properties, particle temperatures are transferred to the grid (P2G), followed by thermal boundary updates and temperature diffusion on the grid solved via a Lattice Boltzmann Method (LBM). 
The updated grid temperature is then interpolated back to the particles (G2P). 
If the particle temperature exceeds a predefined melting threshold $T_{melt}$, we switch its material phase and update the constitutive model accordingly.
For the solid state, we adopt a StVK elasticity model~\cite{klar2016drucker}, while the melted state can be treated as a viscoplastic material and we model it using a Herschel–Bulkley plasticity formulation~\cite{yue2015continuum}.
The thermal boundaries are user-defined heat source regions that 
directly control the spatial extent and intensity of melting, as shown in Fig.~\ref{fig:our_process}.
The overall thermomechanical pipeline is summarized in Algorithm~\ref{alg:mpm} and 
an expanded version with details is provided in \textbf{supplementary material}.
\\
\noindent\textbf{Mapping Deformation to 3D Gaussians.}
After each simulation step $t$, following \cite{xie2024physgaussian}, 
the elastic deformation gradient $\mathbf{F}_{p}^{E,t}$ for splat $G_p$ updates the covariance via an affine push-forward:
\begin{equation}
    \boldsymbol{\Sigma}_p^{t}
    = \mathbf{F}_{p}^{E,t} \, \boldsymbol{\Sigma}_p^{0} \, (\mathbf{F}_{p}^{E,t})^{\top}.
    \label{eq:gs-cov-update}
\end{equation}
Then we decompose $\boldsymbol{\Sigma}_p^{t}=\mathbf{R}_{p}^{t}(\mathbf{S}_{p}^{t}{\mathbf{S}_{p}^{t}}^T){\mathbf{R}_{p}^{t}}^T$ to obtain the updated rotation matrix $\mathbf{R}_{p}^{t}$ and scaling matrix $\mathbf{S}_{p}^{t}$,
and the orientation-dependent appearance is updated by the relative rotation $\Delta \mathbf{R}_p^{t}
= \mathbf{R}_p^{t}{\mathbf{R}_p^{0}}^{T}$: we rotate the view direction $d$ for SH color as:
$c_p^{t}(\mathbf{d})
    = c_p^{0} \bigl( (\Delta\mathbf{R}_{p}^{t})^{\top} \mathbf{d} \bigr)$.
}
\begin{figure}[t] 
    \centering
    \scriptsize
    \includegraphics[width=0.95\textwidth]{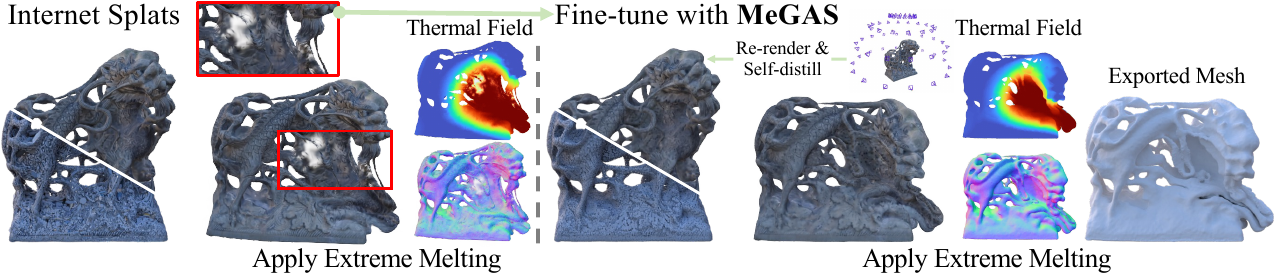}
    \caption{
    \textbf{Naïve combination fails under extreme topology changes.}
    Our thermomechanical augmentation can directly apply to Internet-pretrained splats, nevertheless, naïve 3DGS cracks under large deformations. With low-cost self fine-tuning (1000 steps $\approx$ 10s), \textbf{MeGAS} makes pretrained splats robust and yields smoother geometry.
    }
    \label{fig:extreme_melt_motivation}
\end{figure}

\vspace{-1.0 em}
\subsection{Topology-Adaptive Gaussian Rendering}
\label{sec:method:rendering}

After obtaining particle-level melting animation through our thermomechanical MPM-LBM simulation, 
we achieve a unified framework for elastoplastic simulation and photorealistic rendering. 
However, since melting entails extreme non-rigid deformation and topology change (e.g., surface cracking, internal exposure), naïvely applying PhysGaussian under such conditions leads to severe visual artifacts, e.g., hollow surfaces and needle-like internal floaters, as shown in Fig.~\ref{fig:extreme_melt_motivation}. 
To address these issues, we propose a topology-adaptive 3D Gaussian Splatting (3DGS) framework specifically designed for stable rendering under large-scale melting deformation with the following innovations.

\noindent\textbf{Internal-Free Uniform Gaussian Regularization.}
The original 3D Gaussian Splatting (3DGS) reconstructs object details through outside-in multi-view photometric supervision, which leads to undesirable internal distribution of redundant Gaussians, and consequently degrades rendering quality when hidden Gaussians are exposed under large deformation. Additionally, the reconstructed surface Gaussians typically exhibit highly irregular anisotropy and non-uniform spatial distribution, resulting in needle-like visual artifacts during melting. 

We build upon the planar-based Gaussian Splatting~\cite{huang20242d} and further enhance the training stage. 
Previous work~\cite{barron2022mipnerf360} observes that enforcing a delta-like per-pixel alpha weight distribution along each ray effectively suppresses floating artifacts in volumetric rendering. 
Inspired by this observation, since 3DGS employs an explicit point-based representation, we propose a direct yet effective strategy—pruning low-contribution Gaussians during training.
Specifically, we traverse all training views and record the maximum weight $w_i^{\text{max}}$ of each Gaussian $G_i$ across all pixels. 
Then we prune Gaussians whose $w_i^{\text{max}}$ falls below a predefined threshold $\tau_{\text{prune}}$ which contribute negligibly to the rendered appearance:
$\mathcal{G}' = \{\, G_i \in \mathcal{G} \mid w_i^{\text{max}} > \tau_{\text{prune}} \,\}.$
The remaining subset $\mathcal{G}'$ is subsequently fine-tuned until convergence, eliminating redundant internal Gaussians and continuing the original 3DGS training pipeline.

Irregular anisotropic Gaussians often lead to needle-like kernels or large volumetric discrepancies, which deteriorate rendering stability under large deformation. 
Following~\cite{xie2024physgaussian,feng2025gaussian}, we adopt an anisotropy loss that penalizes excessive scale ratios and encourages the kernel to approach a disk-like shape:
\begin{equation}
\mathcal{L}_{\text{aniso}} = \frac{1}{|\mathcal{P}|} 
\sum_{p \in \mathcal{P}} 
\max \left( 
\frac{S_p^{1}}{S_p^{2}} - a, \, 0 
\right),
\end{equation}
where $a$ is a ratio threshold, and $\mathbf{S}_p = \{S_p^{1}, S_p^{2}, S_p^{3}\}$ denotes the principal scalings of Gaussian $G_p$, 
with $S_p^{1}$ being the largest and $S_p^{3}$ the smallest scaling factor.

We further apply scale clamping with a threshold $\tau_{\text{scale}}$ to prevent large Gaussian volumetric discrepancies: 
$S_p\leftarrow\min(S_p, \, \tau_{\text{scale}}).$  
Densification in 3DGS naturally compensates for potential sparse regions caused by the clamping, 
yielding a uniform, volume-consistent, and surface-aligned Gaussian distribution.
As illustrated in Fig.~\ref{fig:exp_deform}, our regularization produces a uniform Gaussian distribution and reduces internal floaters compared to the regularization used in  PhysGaussian~\cite{xie2024physgaussian} and VR-DoH~\cite{luo2025vrdoh} without compromising rendering fidelity.

\begin{wrapfigure}{r}{0.4375\textwidth}
    \vspace{-2.0 em} 
    \centering
    \scriptsize
    \includegraphics[width=\linewidth]{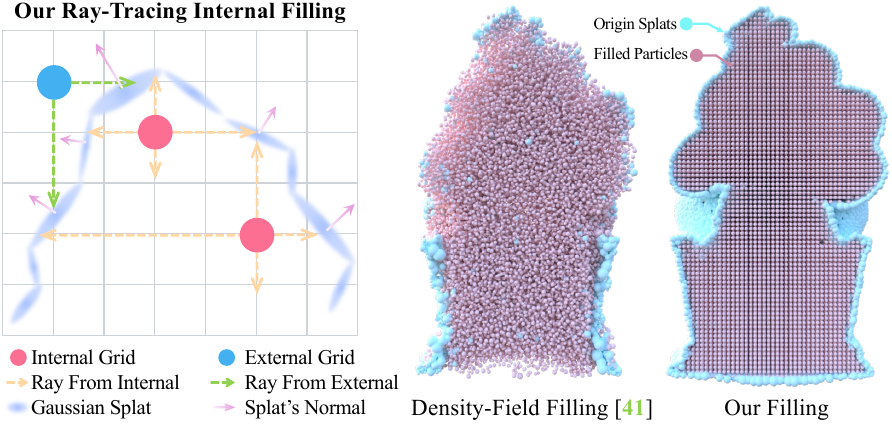}
    \caption{\textbf{Our ray-tracing internal filling.}
    Starting from an interior-free, uniformly distributed GS model, we cast rays from uniform grids and detect interior samples by enforcing directional consistency with Gaussian ray-traced normals. This produces uniform, well-conditioned volumetric filling for physically plausible melting.}
    \label{fig:internal_filling}
    \vspace{-1.25 em} 
\end{wrapfigure}

\begin{figure}[t] 
    \centering
    \scriptsize
    \includegraphics[width=0.95\textwidth]{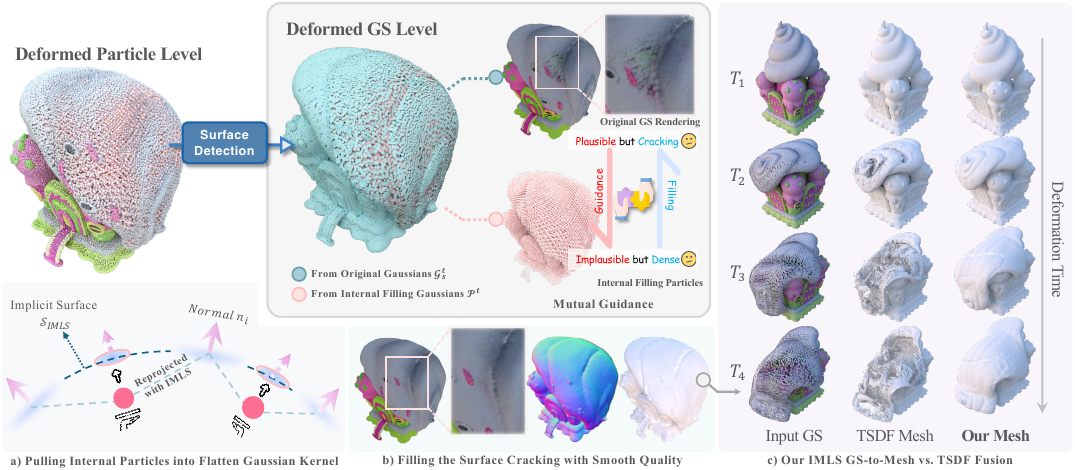}
    \caption{
    \textbf{Implicit-surface–guided adaptive densification.} 
    Internal particles migrating to cracks are detected and projected onto an IMLS-fitted implicit surface to reorient Gaussian kernels with surface-aligned normals.
    Locality-aware densification seals emerging cracks and maintains coherent splat distributions under severe thermomechanical topology change.
    Integrating Gaussian Splatting with IMLS exports temporally more coherent and smoother surfaces while preserving fine geometric details.
    }
    \label{fig:imls_guidance}
    \vspace{-0.5 em}
\end{figure}

\noindent\textbf{Ray-Tracing-Based Internal Filling}.
To enable a physically consistent melting simulation, the object interior needs to be properly filled for volumetric support.
Since our reconstruction is internal-floater-free, we adopt a simple yet robust Gaussian-ray-tracing–based strategy~\cite{3dgrt2024} to detect interior regions.
For each sampled ray, if the accumulated alpha mask indicates full opacity and the blended normal aligns with the ray direction, the ray is classified as interior-originated.
As shown in Fig.~\ref{fig:internal_filling}, we divide the object's bounding box into uniform grids, each grid casts rays along principal directions. 
Grids with a majority of interior-originated rays are supplemented into internal particles.
\\
\noindent\textbf{Adaptive Surface Cracking Densification.}
Melting induces extreme non-rigid deformation and frequent topology changes, which often cause surface cracking and hollow artifacts, as shown in Fig.~\ref{fig:imls_guidance}.  
We observe that the internal particles filled naturally migrate toward surface voids during the MPM evolution, partially compensating for the missing geometry.  
However, these internal particles lack valid Gaussian kernel parameters (covariance/appearance), thus cannot be rendered coherently.
Let the initial surface Gaussians be denoted as $\mathcal{G}_s = \{ G_i^s = (\mathbf{x}_i^s, \mathbf{R}_i^s, \mathbf{S}_i^s, \mathbf{o}_i^s)\}$.  
After deformation step $t$, the internal filled particles $\mathcal{P}_t = \{\mathbf{p}_j^t\}$ geometrically occupy the cracked regions of $\mathcal{G}_s$ but without valid kernel parameters.  
To reconstruct a coherent surface representation, we leverage the deformed surface Gaussians $\mathcal{G}_s^t=\Phi_t(\mathcal{G}_s)$ to guide the local Gaussian kernel fitting of nearby internal particles.
\\
\noindent\textbf{Surface Detection.}
We first identify which internal particles are suitable for use in cracking filling with surface detection.
Specifically, we place virtual cameras uniformly distributed over $360^\circ$.  
Internal particles $\mathbf{p}_j^t$ are initialized as isotropic Gaussian splats $G_j^p$ with opacity $\mathbf{o}_j=1$ and radius $r_j$:
\begin{equation}
r_j=\kappa\,\min\!\left\{\,\min_i\|\mathbf{p}_j^t-\mathbf{x}_i^s\|,\;\min_{k\neq j}\|\mathbf{p}_j^t-\mathbf{p}_k^t\|\,\right\},
\end{equation}
where $\kappa\in(0,1)$ is a shrinkage ratio. 
For each pixel, we record the Gaussian splat $G_i$ with maximal alpha contribution along the ray, 
and yields the surface subset $\{\widetilde{\mathcal{G}}_s^t, \;
\widetilde{\mathcal{P}}^t\}$.

\noindent\textbf{Implicit-Surface Guidance}.
As shown in Fig.~\ref{fig:imls_guidance}, to restore plausible Gaussian kernel for $\widetilde{\mathcal{P}}^t$, we fit an implicit surface $\mathcal{S}_{\text{IMLS}}$ from $\widetilde{\mathcal{G}}_s^t$ via Implicit Moving Least Squares (IMLS)~\cite{alexa2001point,kolluri2008provably,oztireli2009feature}:
\begin{equation}
f(\mathbf{x}) = \frac{\sum_i \phi(\|\mathbf{x}-\mathbf{\widetilde{x}}_i^s\|)\, (\mathbf{\widetilde{n}}_i^s)^\top(\mathbf{x}-\mathbf{\widetilde{x}}_i^s)}
{\sum_i \phi(\|\mathbf{x}-\mathbf{\widetilde{x}}_i^s\|)},
\end{equation}
where $\phi(\cdot)$ is a Gaussian weighting kernel and $\mathbf{\widetilde{n}}_i^s$ is the normal derived from 
$\widetilde{G}_i^s\subset\widetilde{\mathcal{G}}_s^t$. 
Each internal particle is then projected onto $\mathcal{S}_{\text{IMLS}}$ to obtain its surface normal $\mathbf{n}_j$ and rotation matrix $\mathbf{R}_j$ , 
and we further assign color $\mathbf{c}_j$ as the average of $K$ nearest Gaussians in $\widetilde{\mathcal{G}}_s^t$,
yielding a surface-aligned kernel:
\begin{equation}
G_j^p=\bigl(\mathbf{p}_j^t,\mathbf{R}_j,\mathbf{S}_j,\mathbf{o}_j,\mathbf{c}_j\bigr),
\; \text{where} \; \mathbf{S}_j=r_j\mathbf{I}.
\end{equation}
This adaptive densification seamlessly re-covers cracked regions with correctly oriented internal Gaussians and yields smooth, stable rendering under extreme melting deformation. 
Meanwhile, the fitted implicit surface $\mathcal{S}_{\text{IMLS}}$ also enables direct extraction of high-quality meshes.
See more details in \textbf{supplementary}.

\vspace{-1.5 em}
\section{Experiments}
\label{sec:experiment}
\vspace{-0.25 em}

We evaluate our \method on diverse synthetic and real-world scenes to demonstrate our capability to produce physically plausible, photorealistic thermophysical editing and robustness under extreme deformations.

\vspace{-1.5 em}
\subsection{Evaluation on Thermophysical Scene Editing}
\vspace{-0.5 em}

\begin{figure*}[t] 
    \centering
    \scriptsize
    \includegraphics[width=0.975\textwidth]{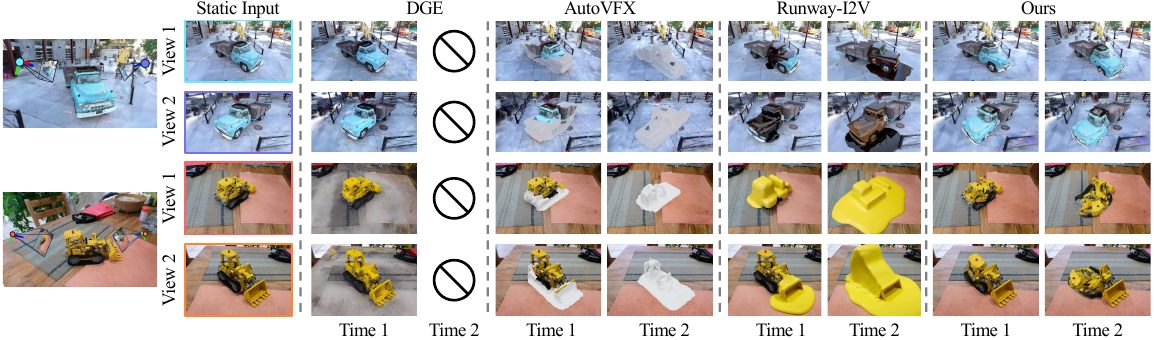}
    \caption{
    \textbf{Melting-style scene editing comparisons.}
    Qualitative results with multi-view, temporally evolving melting edits across DGE, AutoVFX, Runway Gen-4.5, and our \method show that prior methods lack physical realism, photorealism, or multi-view consistency, whereas \method preserves photorealistic appearance and delivers physically plausible, temporally consistent melting. 
    \textbf{Please refer to our supp. video for more vivid dynamic comparisons.}
    }
    \label{fig:exp_edit}
\end{figure*}

\noindent\textbf{Datasets and Comparison.}
To evaluate \method as a practical editing tool, we perform melting-style edits on in-the-wild scenes from MipNeRF360~\cite{barron2022mipnerf360} and Tanks and Temples~\cite{knapitsch2017tanks} datasets.
For each scene, we first obtain a high-quality static 3DGS reconstruction, and then compare our melting edits against three representative state-of-the-art 
approaches:
1) DGE~\cite{chen2024dge}, a text-driven diffusion-based editing method for 3DGS; 
2) AutoVFX~\cite{hsu2025autovfx}, which leverages an LLM~\cite{achiam2023gpt} to generate scripts for external physics engines to perform mesh-based melting;
3) Runway Gen-4.5~\cite{runway_gen4}, a leading commercial image-to-video generation model that augments a single input image with text-guided dynamic effects. 
\textbf{Notably}, existing physics-integrated neural rendering methods do not support thermophysical animation or phase-change-aware dynamics, therefore, we compare against these general editing baselines (including a commercial open-domain video generator) to contextualize our thermophysical editing.
All methods are provided with comparable editing instructions, and detailed prompts and parameter settings are reported in \textbf{supplementary material}.



\noindent\textbf{Results.}
Qualitative comparisons are shown in Fig.~\ref{fig:exp_edit}. 
DGE fails to induce coherent dynamic melting behavior and exhibits limited physical realism.
AutoVFX generates plausible mesh-based melting, 
but decoupled geometry/shading from radiance causes non-photorealistic compositions and inconsistencies with the original scene.
Runway Gen-4.5 produces visually vivid effects, yet the motion is often physically implausible, object identities drift over time, and it lacks multi-view consistency.
These limitations stem from the absence of explicit physical modeling, which makes it difficult to capture complex physical phenomena.
In contrast, by integrating thermomechanical simulation with neural scene representations, \method preserves the original photorealistic appearance and produces multi-view consistent, physically plausible thermophysical dynamics.
\textbf{
Please refer to supp. for more edited results on real-world data.
}
\\
\vspace{-1.0 em}

\begin{wraptable}{r}{0.55 \textwidth}
\vspace{-2.25 em} 
\captionsetup{font=footnotesize}
\caption{
\textbf{User Study.} 
The rankings assigned by participants to each method are converted to integer scores (best=4, worst=1) and averaged over all videos and raters.
}

\vspace{0.75 em}
\resizebox{\linewidth}{!}{%
\begin{tabular}{ccccc}
\specialrule{.1em}{.1em}{.1em}
Methods & DGE~\cite{chen2024dge} & AutoVFX~\cite{hsu2025autovfx} & Gen-4.5~\cite{runway_gen4} & Ours \\
\toprule
Physical-Realism $\uparrow$        & 1.538 & 2.692 & 2.038 & \textbf{3.731} \\
Photo-Realism $\uparrow$           & 2.038 & 2.154 & 2.154 & \textbf{3.654} \\
Multi-View Consistency $\uparrow$  & 2.154 & 2.385 & 1.885 & \textbf{3.577} \\
\specialrule{.1em}{.1em}{.1em}
\vspace{-5.0 em}
\end{tabular}

}

\label{tab:user_study}
\end{wraptable}
\noindent\textbf{User Study.}
We further conduct a user study as shown in Tab.~\ref{tab:user_study} across physical plausibility, photorealism, and multi-view/temporal consistency, supporting our thermomechanical dynamic Gaussian Splatting modeling yields more convincing and reliable melting edits (see \textbf{supp.} for full config.).
\subsection{Evaluation on Extreme Melting Deformation}

\begin{figure*}[t] 
    \centering
    \scriptsize
    \vspace{-0.75 em}
    \includegraphics[width=\textwidth]{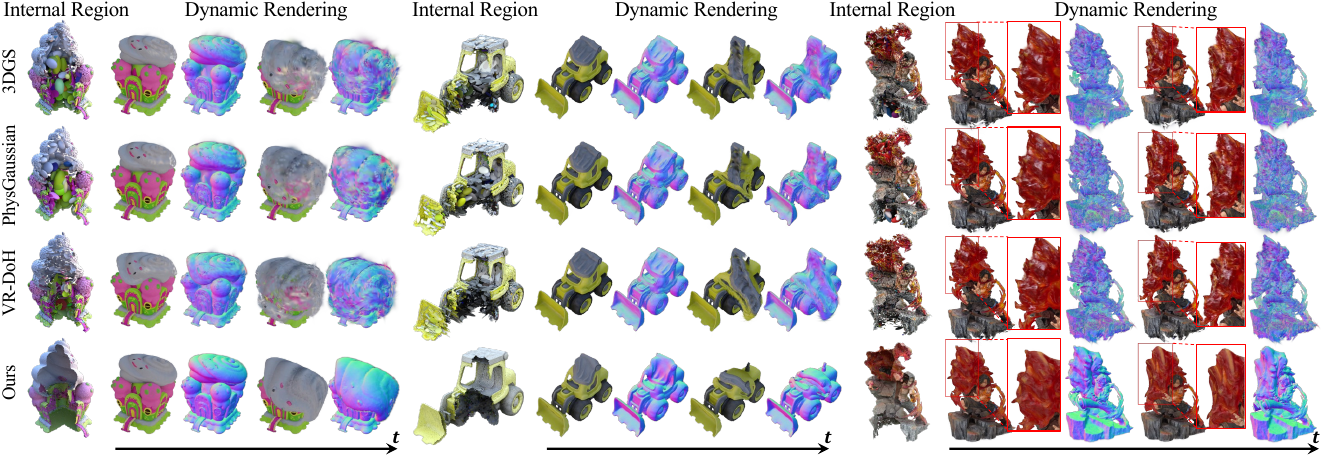}
    \caption{
    \textbf{Extreme melting deformation comparisons.}
    Internal cross-sections and dynamic frames show that Original 3DGS, PhysGaussian, and VR-DoH accumulate internal floaters, needle-like artifacts, and surface cracking under large deformations, whereas \method yields volume-consistent, crack-free geometry and stable rendering throughout melting.
    \textbf{Please refer to supp. video for vivid dynamic comparison.}
    }
    \label{fig:exp_deform}
\end{figure*}

\noindent\textbf{Datasets and Comparison.}
We evaluate rendering quality under extreme thermomechanical deformations with a set of Blender-synthetic objects and real-world scenes captured with an iPhone. We compare the following 3DGS-based deformation pipelines: 
1) Original 3DGS~\cite{kerbl3Dgaussians}; 
2) PhysGaussian~\cite{xie2024physgaussian}, a physics-integrated 3DGS with isotropic constrained kernels; 
3) VR-DoH~\cite{luo2025vrdoh}, which enforces volume-consistency constraints.
All baselines are coupled with the same MPM-LBM simulator, and differ only in how Gaussians are regularized, internally filled, or adapted to deformation.

\begin{figure}[t] 
    \centering
    \scriptsize
    \vspace{-0.75 em}
    \includegraphics[width=0.95\textwidth]{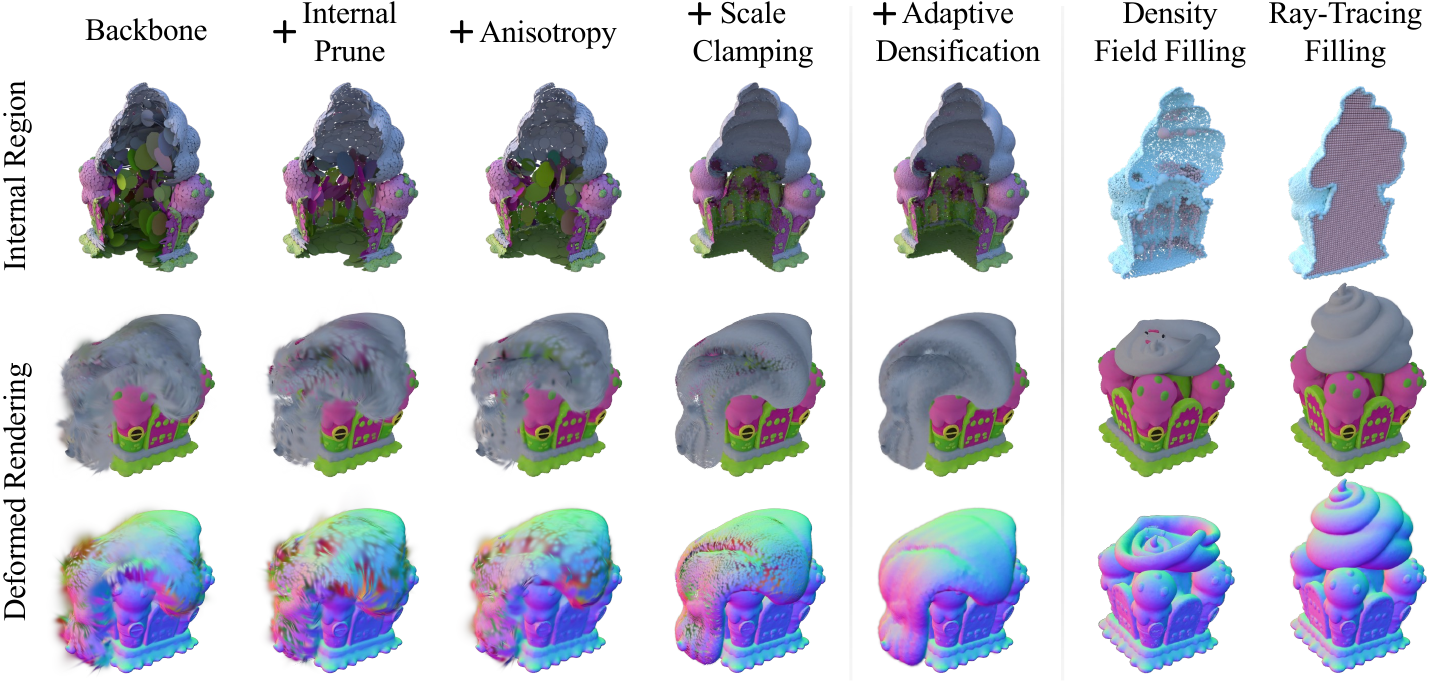}
    \caption{
    \textbf{Ablations on Topology-Adaptive strategy.} 
    Our topology-adaptive strategy yields an interior-free, uniformly distributed Gaussian representation with consistent volume, mitigating needle-like artifacts and internal floaters. 
    The adaptive densification fills surface cracking during animation, and our ray-tracing–based internal filling uniformly populates the interior, providing robust volumetric support.
    }
    \label{fig:ablation_regular}
\end{figure}

\noindent\textbf{Implementation Details.}
For each scene, we first reconstruct 3DGS with the respective methods, adopting the internal filling strategies proposed for each baseline (e.g., density-field–based schemes~\cite{xie2024physgaussian}) and employing our ray-tracing–based filling strategy for \method.
We then manually place identical heat sources across methods, and evolve the thermal field using the same LBM-based advection–diffusion solver with consistent thermal diffusivity, boundary conditions, and phase-transition parameters.
The resulting thermal field drives MPM particles according to phase-aware constitutive models, and Gaussians are updated at each step via deformation-gradient–based transformations.
All methods are simulated for the same duration with identical material and solver settings (\textbf{see supp. for full configurations and computational cost}).

\noindent\textbf{Results.}
Fig.~\ref{fig:exp_deform} visualizes internal cross-sections and dynamic frames under large-scale melting.
Baseline methods accumulate internal floaters, produce severe needle-like artifacts, and suffer from surface cracking that exposes underspecified interiors.
In contrast, our \method yields an internal-floater-free and volume-consistent Gaussian distribution, with stable, crack-free surfaces refined by our adaptive surface densification. The resulting geometry remains smooth and coherent throughout the melting process, allowing direct extraction of high-quality meshes from the evolving Gaussian representation, as illustrated in Fig.~\ref{fig:our_process}.

\subsection{Ablation Studies}

\noindent\textbf{Topology-Adaptive Strategy.}
As shown in Fig.~\ref{fig:ablation_regular}, we progressively incorporate our topology-adaptive strategy into the backbone model~\cite{chen2024pgsr}. 
Internal-Prune 
effectively suppresses internal floaters,
while anisotropy loss and scale clamping yield a uniform, volume-consistent Gaussian distribution that avoids needle-like artifacts under large deformation.
During animation, adaptive densification compensates for surface cracking and maintains smooth rendering.
Since our representation is interior-free, directly applying density-field–based filling~\cite{xie2024physgaussian} leaves the interior hollow and prone to collapse, whereas our ray-tracing filling uniformly populates the volume and provides robust volumetric support.

\begin{figure}[t]
    \centering

    
    \begin{minipage}[t]{0.55\linewidth}
        \vspace{0pt} 
        \centering
        \scriptsize
        \vspace{-0.75 em}
        \includegraphics[width=\linewidth]{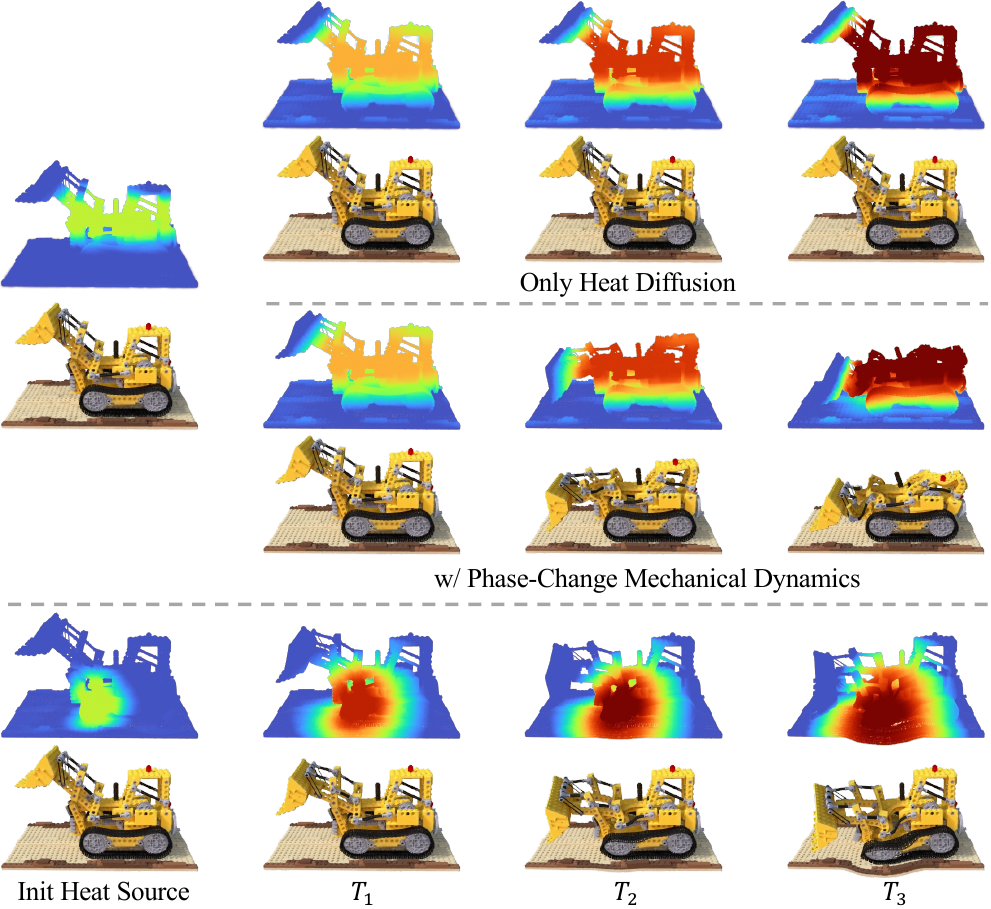}
        \captionsetup{font=footnotesize}
        \captionof{figure}{
        \textbf{Heat diffusion with constitutive \\ model switching.}
        Phase-change-driven constitutive switching couples melting to temperature evolution, while different heat-source configurations yield distinct melting patterns.
        }
        \label{fig:diffusion_process}
    \end{minipage}
    \begin{minipage}[t]{0.425\linewidth}
        \vspace{0pt} 
        \centering
        \scriptsize
        \vspace{-0.75 em}
        \includegraphics[width=\linewidth]{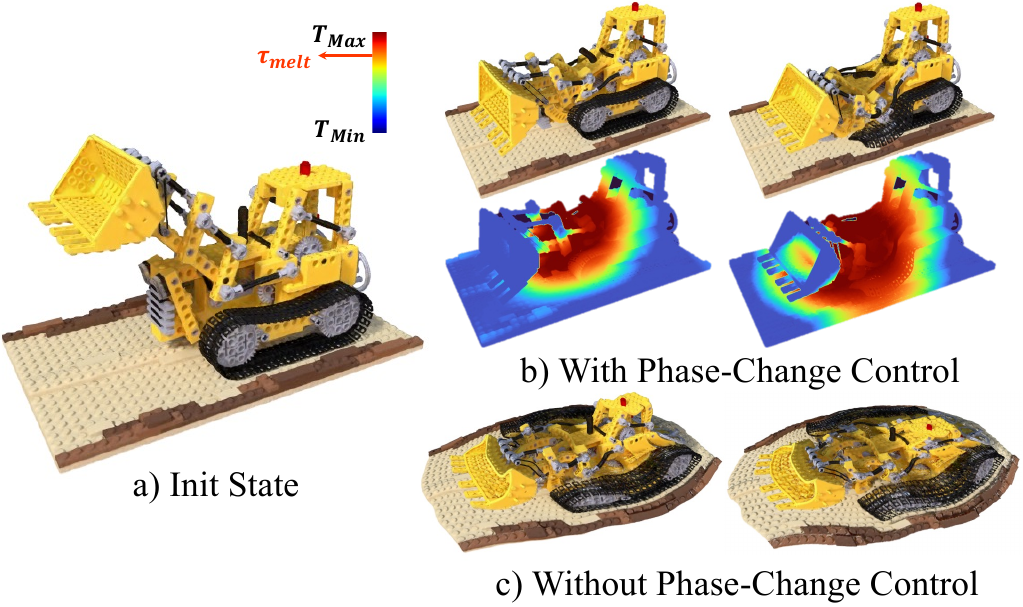}
        \captionsetup{font=footnotesize}
        \captionof{figure}{
        \textbf{Ablations on phase-change switching.}
        Without phase change, melting is triggered globally and collapses prematurely. 
        }
        \label{fig:ablation_phase}

        \vspace{1.0 em}

        \includegraphics[width=\linewidth]{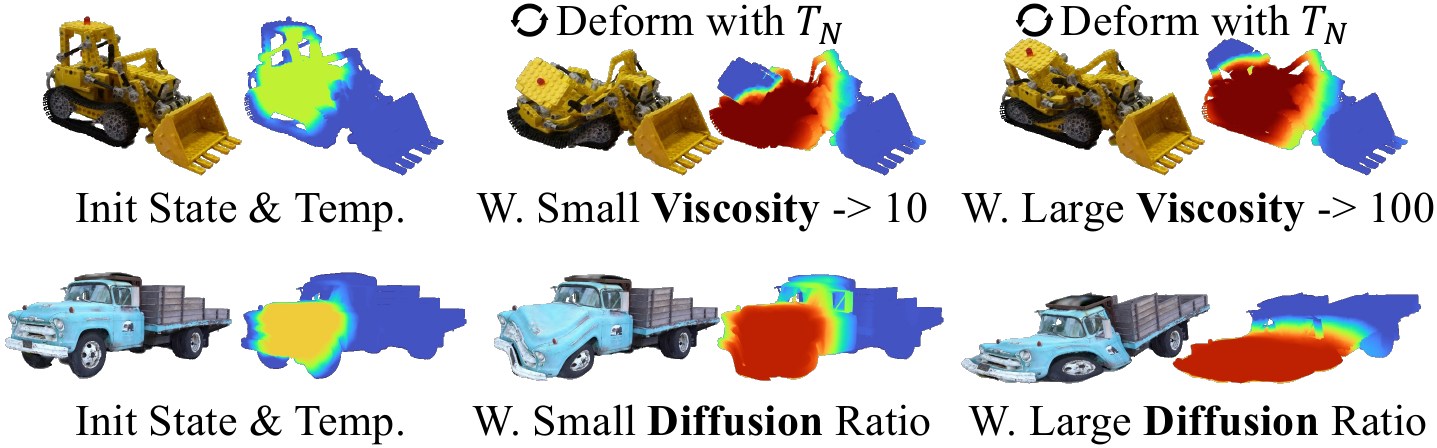}
        \captionsetup{font=footnotesize}
        \captionof{figure}{
        \textbf{Controllable melting via parameter adjustment.}
        By modifying thermophysical parameters, users controllably achieve different melting effects.}
        \label{fig:ablation_controllable}
        \vspace{0.5 em}
    \end{minipage}
\end{figure}

\noindent\textbf{Phase-Change Switching.}
As shown in Fig.~\ref{fig:ablation_phase}, we demonstrate the necessity of phase-change–driven constitutive switching by comparing an init-melted strategy, where all Gaussians are initialized in the melted state and follow the viscoplastic constitutive model, while we incrementally change each Gaussian’s phase state based on the evolving temperature field.
Init-melted strategy exhibits premature global slumping, while our temperature-controlled transition yields a more physically plausible melting progression.
Moreover, we present visualizations of pure heat diffusion in Fig.~\ref{fig:diffusion_process}, where the temperature field progressively propagates from high- to low-temperature regions over time. 
Building on this diffusion process, our phase-change-aware constitutive switching produces melting dynamics explicitly coupled to temperature evolution. 
We additionally show that interactive heat-source specification (Fig.~\ref{fig:diffusion_process}) and thermophysical parameter tuning (Fig.~\ref{fig:ablation_controllable}) enable controllable melting with distinct patterns and effects.

\subsection{Scalability to Complex Thermomechanical Interactions.}

\begin{figure}[t] 
    \centering
    \scriptsize
    \vspace{-0.75 em}
    \includegraphics[width=\textwidth]{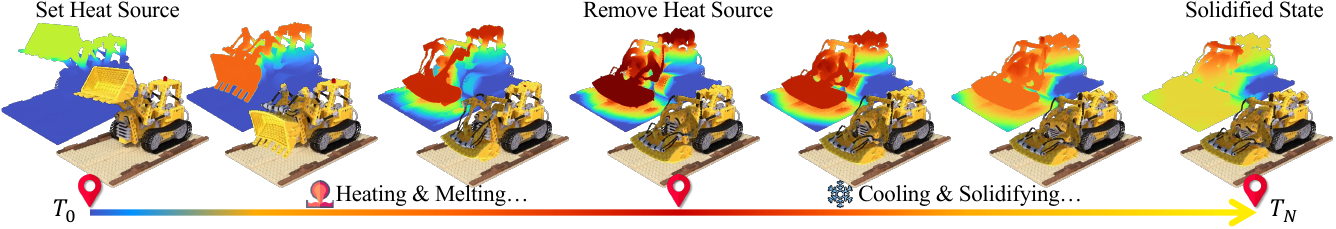}
    \caption{
    \textbf{Bidirectional phase-change editing.}
    Controlled heating induces melting, while subsequent cooling drives solidification, demonstrating consistent thermomechanical evolution across phase transitions and enabling general thermophysical editing.
    }
    \vspace{0.75 em}
    \label{fig:melt2solid}
\end{figure}

Built upon an MPM-based simulator, our framework supports broader thermomechanical interactions beyond single-object melting. In addition, the cooling-driven solidification shown in Fig.~\ref{fig:melt2solid} demonstrates that our framework naturally supports bidirectional phase transitions, extending thermomechanical editing beyond melting. As illustrated in Fig.~\ref{fig:complex_scene}, we further showcase collisions with real-scene environments, interactions with complex geometric colliders and elastic bodies, and heat-transfer interactions among multiple melting objects. 
Our method preserves the inherent extensibility of MPM while maintaining stable rendering under severe deformations, thereby broadening dynamic neural rendering toward richer thermomechanical dynamics modeling.

\begin{figure}[t]
    \centering
    \scriptsize
    \animategraphics[poster=20,autoplay,loop,width=\linewidth]{15}{figures/complex_scene/}{000000}{000039}
    \caption{
    \textbf{Scalability to complex scenes.} 
    Our method naturally supports multi-object interactions with multi-material and collisions (e.g., an ice cream ball). \textbf{Please view the dynamic videos in \textcolor{red}{Adobe Acrobat Reader}}!
    }
    \label{fig:complex_scene}
\end{figure}

\section{Conclusion}
\label{sec:conclusion}
We have proposed \method, a novel framework that integrates thermomechanical dynamics into 3D Gaussian Splatting. 
\method jointly models appearance and a controllable thermal field, coupling a heat advection-diffusion solver with temperature-conditioned constitutive switching to realize phase-change–driven thermophysical editing. 
To handle extreme deformations that induce topology changes and surface cracking, we combine training-time anisotropy regularization and interior-free filtering with an online deformation-stage refinement that we fit local implicit surfaces to seal cracks and yield smooth renderings. To our knowledge, \method provides the first exploration of thermomechanical phase-change dynamics within a neural rendering pipeline, addressing an underexplored yet fundamental dimension of physics integration.
\\
\noindent\textbf{Limitation.}
We currently do not model material-dependent appearance changes induced by phase transitions (e.g., PBR property variations during melting), as jointly coupling thermomechanics with dynamic material appearance remains highly challenging. 
A promising direction is to 
integrate our framework with video diffusion models~\cite{liu2024physgen, wan2025wan, blattmann2023stable} to enhance the final appearance rendering.

\clearpage
\section*{Acknowledgments}
This work was partially supported by the National Natural Science Foundation of China (NSFC) under Grants 62572425 and 624B2132.
We thank the anonymous reviewers for their valuable comments and suggestions.

%
%
\bibliographystyle{splncs04}
\bibliography{main}

\end{document}